\documentclass[]{bytedance_seed}

\usepackage{float}
\usepackage{tikz}

\title{REER-PT: Reverse-Engineered Reasoning for Perplexity-Guided Pre-training Data Augmentation}

\author[1,2,\dagger,*]{Haoran Que}
\author[2]{Jiajun Shi}
\author[2]{Ting Huang}
\author[2]{Renming Pang}
\author[3,4]{Jiaheng Liu}
\author[\ddagger,\dagger]{Ge Zhang}
\author[2]{Wenhao Huang}
\author[2]{Shen Yan}
\author[1, \dagger]{Wei Ye}
\author[1]{Shikun Zhang}

\affiliation[1]{Peking University}
\affiliation[2]{ByteDance Seed}
\affiliation[3]{Nanjing University}
\affiliation[4]{TokenWave.AI}

\contribution[*]{Work done at ByteDance Seed}
\contribution[\dagger]{Corresponding authors}
\contribution[\ddagger]{Independent Researcher}

\abstract{
As language-model compute continues to scale, high-quality training data is becoming an increasingly important bottleneck. Conventional next-token prediction supervises what follows a context but leaves the intermediate reasoning behind that continuation implicit. We introduce \textbf{REER-PT}, a scalable framework that extends Reverse-Engineered Reasoning (\textbf{REER}) to raw pre-training data. REER-PT identifies continuations that are difficult to predict but can still be inferred from the preceding context, and inserts concise reasoning annotations that reconstruct the missing connection between context and continuation. Candidate annotations are generated and refined offline, with perplexity serving as the optimization signal. Constraints on length and target leakage filter out unhelpful or trivial annotations. This sparse transformation preserves the source text and remains compatible with standard next-token prediction, avoiding online reasoning rollouts during pre-training. We apply REER-PT to transform a source pre-training corpus into an augmented one. Across augmented-data, original-token, and selected-continuation comparisons, perplexity reductions range from 0.42 to 7.29, and only about 0.05\% of annotation 13-grams appear verbatim in the source text.
We then train two 680M-parameter models with the same architecture and training configuration on the source and augmented corpora, respectively. The augmented-data model gains up to 2.07 percentage points on several knowledge and reasoning benchmarks. Together, the perplexity analysis indicates improved continuation predictability, while the controlled pre-training experiments suggest that this augmentation can improve model performance without changing the standard pre-training objective.
}

\date{\today}
\correspondence{Haoran Que at \email{hrque25@stu.pku.edu.cn}, Ge Zhang at \email{gezhang@umich.edu}, Wei Ye at \email{wye@pku.edu.cn}}

\begin{document}
\maketitle
\begin{tikzpicture}[remember picture,overlay]
    \node[anchor=north east,inner sep=0pt] at ([xshift=-2.4cm,yshift=-1.8cm]current page.north east) {\includegraphics[width=4.3cm]{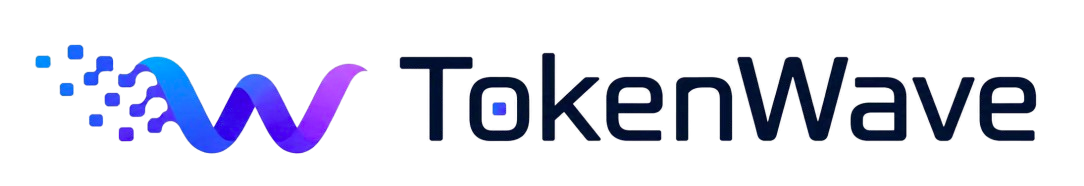}};
\end{tikzpicture}

\section{Introduction}
\label{sec:introduction}

Large language models acquire most of their capabilities through next-token prediction on massive text corpora \citep{kaplan2020scaling,brown2020language}. As compute continues to scale, however, high-quality training data is becoming an increasingly important bottleneck \citep{hoffmann2022training,lee2021deduplicating,gunasekar2023textbooks}. Conventional pre-training teaches a model what text follows a context, but rarely explains why it follows. We therefore consider explicitly annotating the intermediate reasoning that connects a context to its continuation. Chain-of-thought (CoT) supervision provides a natural mechanism for representing these dependencies explicitly \citep{wei2022chain,zelikman2022star}, allowing reasoning signals to be incorporated directly into the training text \citep{zelikman2024quietstar,wang2025tpt}. Yet most conventional CoT datasets are built from curated question--answer pairs rather than ordinary documents, limiting their scale and domain coverage. In contrast, pre-training corpora already span diverse subjects, genres, and forms of discourse. Synthesizing reasoning annotations directly from these corpora therefore offers a scalable way to enrich existing data and provide reasoning signals across domains without designing separate tasks for each domain \citep{maini2024rephrasing}.

Recent work has introduced reasoning signals into pre-training data, either through generated or latent intermediate thoughts \citep{zelikman2024quietstar,ishibashi2025mining,ruan2025latent,wang2025tpt} or through reinforcement learning on observed continuations \citep{dong2025rpt,hatamizadeh2025rlp,li2025rlpt}. These directions establish the promise of reasoning-aware pre-training, but leave two practical challenges. First, dense or online reasoning generation can be expensive at corpus scale. Second, a fluent model-generated annotation is not necessarily useful. It may be redundant, weakly grounded, or unrelated to the actual difficulty faced by the model. Moreover, high-loss tokens are not always reasoning opportunities; some are unpredictable because they introduce arbitrary names, dates, identifiers, or external facts. Effective corpus augmentation must therefore determine both where reasoning is useful and whether a proposed annotation makes the observed continuation easier to predict.

\begin{figure}[h]
    \centering
    \makebox[\linewidth][c]{\includegraphics[width=\linewidth]{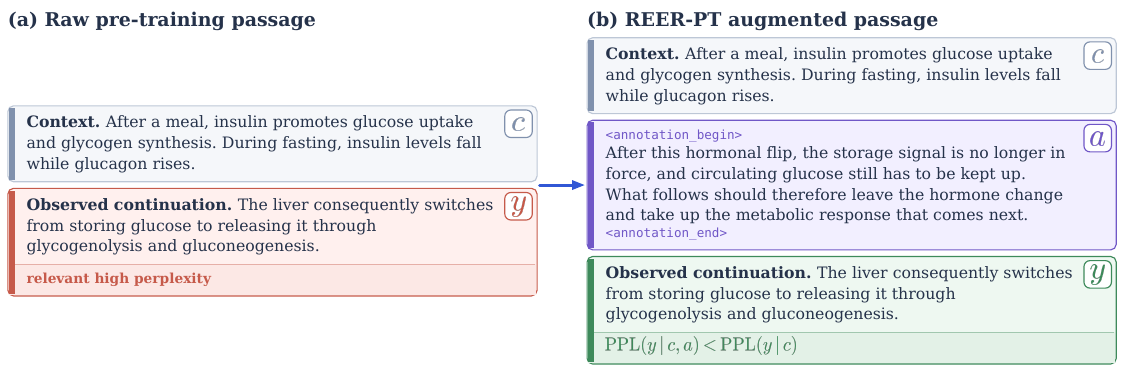}}
    \caption{Illustration of REER-PT augmentation. Given a context $c$ and an observed continuation $y$ with relatively high perplexity, REER-PT inserts a concise reasoning annotation $a$ that makes the implicit transition explicit. The original continuation is preserved, while conditioning on the annotation makes it easier to predict: $\operatorname{PPL}(y \mid c,a) < \operatorname{PPL}(y \mid c)$.}
    \label{fig:reer_pt_example}
\end{figure}

To address these challenges, we introduce \textbf{REER-PT}, a scalable framework that extends Reverse-Engineered Reasoning (REER) \citep{wang2025reer} to pre-training data. REER uses the perplexity of a known reference output as an optimization signal, searching over candidate CoT trajectories for reasoning that makes the reference easier to generate. Following this principle, REER-PT uses the perplexity of observed continuations to guide the search for useful reasoning annotations in raw documents.
It first identifies difficult-to-predict continuations and retains only those that are inferable from context. For each selected continuation, an annotation model produces a concise, book-note-style annotation that reconstructs the missing connection between the preceding context and the continuation. Candidate annotations are then evaluated and refined offline, and only those that reduce continuation perplexity while satisfying constraints on length and target leakage are retained. Restricting this offline search to a sparse set of difficult yet contextually inferable continuations keeps corpus-scale augmentation efficient and selective. Because annotation generation and refinement are completed offline and the source text is preserved, the augmented corpus can be trained with standard next-token prediction without online reasoning rollouts.
Figure~\ref{fig:reer_pt_example} illustrates how a reasoning annotation bridges the context and the observed continuation. In the example, the preceding discussion of insulin and glucagon supports the continuation, but the transition to hepatic metabolism remains implicit. REER-PT inserts a concise, third-person, book-note-style annotation that captures this connection without revealing the target content. Conditioning on the annotation lowers the perplexity of the unchanged continuation.

We apply REER-PT to transform a source pre-training corpus into an augmented one and analyze the resulting data at the augmented-data, original-token, and selected-continuation levels. Across these comparisons, perplexity reductions range from 0.42 to 7.29, and only about 0.05\% of annotation 13-grams appear verbatim in the source text. We then compare two 680M-parameter models with the same architecture and training configuration, trained on the source and augmented corpora, respectively. The augmented-data model gains up to 2.07 percentage points on several knowledge and reasoning benchmarks. Together, the perplexity analysis indicates improved continuation predictability, while the controlled pre-training experiments suggest that this augmentation can improve model performance without changing the standard pre-training objective.

Our main contributions are as follows:

\begin{itemize}
    \item We introduce REER-PT, a scalable framework that augments raw pre-training data with concise reasoning annotations that make implicit connections between context and continuation explicit. It selects difficult continuations that can be inferred from context and retains annotations that reduce continuation perplexity while preserving the original text.

    \item We apply REER-PT at corpus scale and analyze the resulting augmented data at the augmented-data, original-token, and selected-continuation levels. Across these comparisons, perplexity reductions range from 0.42 to 7.29, and only about 0.05\% of annotation 13-grams appear verbatim in the source text. A comparison of two 680M-parameter models with the same architecture and training configuration further shows gains of up to 2.07 percentage points on several knowledge and reasoning benchmarks.
\end{itemize}

\section{Related Work}
\label{sec:related_work}

\paragraph{Pre-training data selection and transformation.}
Existing work improves pre-training data through two broad strategies: selection and transformation. Selection-based methods prioritize specific portions of a corpus. Rho-1 applies selective language modeling, using token-level excess loss relative to a reference model to concentrate optimization on high-excess-loss portions of the corpus \citep{lin2024rho}. The Irreducible Curriculum prioritizes data estimated to have higher learnability, using a small proxy model to approximate loss trajectories \citep{fan2023irreducible}. Transformation-based methods instead rewrite or augment source documents. WRAP rephrases web documents in alternative styles and formats \citep{maini2024rephrasing}, while REWIRE rewrites lower-quality web documents that would otherwise be discarded \citep{nguyen2025rewire}. Active Reading generates document-specific augmentations through self-generated learning strategies to improve factual learning \citep{lin2025activereading}. SwallowCode and SwallowMath rewrite code snippets and mathematical solutions into cleaner, more self-contained forms \citep{fujii2025rewriting}. FineInstructions converts pre-training documents into synthetic instruction--answer pairs using templates derived from real user queries \citep{patel2026fineinstructions}. In contrast, REER-PT performs a targeted local transformation. It inserts a concise reasoning annotation before a selected continuation only if the annotation reduces that continuation's perplexity.

\paragraph{Reasoning-augmented pre-training.}
Recent methods recover or introduce intermediate reasoning in pre-training data.
At the token level, Quiet-STaR \citep{zelikman2024quietstar} generates internal rationales for future-token prediction, adaptive latent CoT \citep{zeng2026adaptivecot} allocates variable-length latent reasoning according to token difficulty, and ToW \citep{xu2024tow} inserts fine-grained explanations for predictable target words.
At the document level, Mining Hidden Thoughts \citep{ishibashi2025mining} reconstructs thought processes underlying STEM and legal texts, Reasoning to Learn from Latent Thoughts \citep{ruan2025latent} infers latent reasoning behind mathematical text, and TPT \citep{wang2025tpt} augments pre-training data with generated thinking trajectories.
Reference-guided methods include REER \citep{wang2025reer}, which uses reference perplexity to search for effective reasoning trajectories, and Understanding by Reconstruction \citep{zeng2026reconstruction}, which recovers planning, reasoning, and debugging trajectories from software repositories.
REER-PT instead inserts concise, book-note-style annotations that explain the connection between the preceding context and a continuation. These annotations resemble concise notes attached to the original document, making them more compatible with pre-training data than long task-oriented or agentic reasoning trajectories.

\paragraph{Reinforcement pre-training and mid-training.}
Recent work also brings reinforcement learning into pre-training and mid-training by deriving rewards or learning signals from pre-training corpora.
In reinforcement pre-training, RPT reframes next-token prediction as a reinforcement-learning task and rewards reasoning that correctly predicts the next token \citep{dong2025rpt}.
RLP treats sampled chains of thought as exploratory actions and rewards them according to the increase in future-token log-likelihood \citep{hatamizadeh2025rlp}.
RLPT extends reinforcement learning from next-token prediction to next-segment prediction, rewarding trajectories that recover subsequent text segments \citep{li2025rlpt}.
PretrainZero trains a reasoning policy to identify informative masked content in a general pre-training corpus and reconstruct it without external labels or verifiers \citep{xing2025pretrainzero}.
At the mid-training stage, RMT combines a dynamic reasoning budget, curriculum-based adaptive sampling, and joint reinforcement and next-token training \citep{tian2025rmt}.
OctoThinker studies how data composition and training schedules during mid-training affect the effectiveness of subsequent reinforcement learning \citep{wang2025octothinker}.
REER-PT, by contrast, searches for annotations offline using continuation perplexity. This avoids policy rollouts and reward optimization during model training. The resulting corpus supports standard next-token prediction and can be constructed at pre-training scale.

\section{Approach}

\subsection{Overview}

We introduce \textbf{REER-PT}, a scalable framework for adding reasoning annotations to raw pre-training data. It extends Reverse-Engineered Reasoning (REER) \citep{wang2025reer} from query--response data to document continuations. Given a source document, REER-PT treats sentence boundaries as candidate insertion positions. At the $i$-th selected position, $c_i$ denotes the local context preceding the boundary, $y_i$ denotes the observed continuation following it, and $a_i$ denotes the final accepted annotation inserted between them. REER-PT uses sentence-level perplexity to locate difficult transitions and continuation perplexity to guide the refinement of a concise, book-note-style annotation $a_i$ that makes the connection from $c_i$ to $y_i$ explicit. Applying this transformation at sparsely selected positions preserves the order and content of all source tokens.

\begin{figure}[h]
    \centering
    \includegraphics[width=\linewidth]{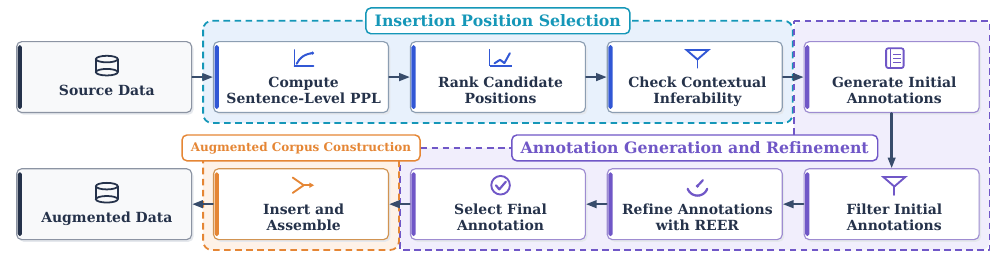}
    \caption{Overview of the REER-PT pipeline. REER-PT ranks candidate insertion positions by sentence-level perplexity, filters them by contextual inferability, generates and refines annotations using continuation perplexity, and inserts the selected annotations into the source documents.}
    \label{fig:reer_pt_pipeline}
\end{figure}

Figure~\ref{fig:reer_pt_pipeline} illustrates the three stages of the REER-PT pipeline: 

First, REER-PT segments each document into sentences, computes sentence-level perplexity, and ranks the sentences from highest to lowest perplexity. High perplexity identifies difficult transitions, but it does not guarantee that a continuation can be inferred from the preceding context. Some sentences instead introduce arbitrary names, dates, identifiers, or external facts. The annotation model therefore performs an inferability check and filters out candidates that cannot be supported by the context. This stage selects insertion positions that are both difficult to predict and contextually inferable.

Second, the annotation model produces multiple initial annotations for each selected position. Before refinement, REER-PT checks each initial annotation to ensure that it falls within a preset length range, typically 500--1,000 words, and does not contain target leakage. Target leakage occurs when an annotation directly repeats or closely paraphrases words or facts from the continuation. Such content can lower continuation perplexity by revealing the target in advance rather than by explaining the connection from context to continuation. REER-PT therefore removes initial annotations that fail either filter. It then splits each remaining annotation into multiple parts and refines each part separately. At each refinement step, REER-PT retains the candidate with the lowest continuation perplexity. After all refinement trajectories are completed, REER-PT selects the refined annotation with the lowest continuation perplexity across all trajectories. If this perplexity is not lower than the no-annotation baseline, REER-PT discards the position without inserting an annotation.

Third, each accepted annotation is delimited by the dedicated boundary markers \texttt{<annotation\_begin>} and \texttt{<annotation\_end>} and inserted immediately before its target continuation. All source tokens remain unchanged and in their original order. The resulting augmented corpus is fixed before pre-training and can be used directly with the standard next-token prediction objective.

\subsection{Pipeline}

\paragraph{Insertion Position Selection.}
Let $D=(x_1,\ldots,x_T)$ denote the token sequence obtained by tokenizing a document, where $T$ is the resulting number of tokens and $x_t$ is the token at position $t\in\{1,\ldots,T\}$. REER-PT involves two model roles. The \textbf{PPL model} provides the token-level log probabilities used to select insertion positions and evaluate annotation candidates; we denote its conditional next-token distribution by $p_{\mathrm{ppl}}$. The \textbf{annotation model} performs the inferability check and generates both initial annotations and candidate rewrites during refinement. The choice of annotation model is flexible. In contrast, the choice of PPL model determines whether data construction is on-policy or off-policy with respect to the target model. Using the target model itself as the PPL model yields on-policy selection and optimization, whereas using a separate PPL model yields an off-policy procedure. Stronger PPL models or checkpoints from different training stages may identify different insertion positions and provide different refinement signals. For each token, we compute
\begin{equation}
\ell_t = \log p_{\mathrm{ppl}}(x_t \mid x_{<t}),
\end{equation}
where $\ell_t$ is the token-level log probability and $x_{<t}=(x_1,\ldots,x_{t-1})$ is the prefix preceding $x_t$. We segment $D$ into $m$ sentences $s_1,\ldots,s_m$, where $s_j$ denotes the $j$-th sentence, and let $I_j$ be the set of token indices belonging to $s_j$. We first compute the average token log probability of each sentence:
\begin{equation}
\bar{\ell}_j = \frac{1}{|I_j|}\sum_{t\in I_j}\ell_t.
\end{equation}
Here, $j\in\{1,\ldots,m\}$, $I_j\subseteq\{1,\ldots,T\}$, $|I_j|$ is the number of tokens in $s_j$, and $\bar{\ell}_j$ is the average token log probability of that sentence. We define sentence-level perplexity as
\begin{equation}
P_j=\exp(-\bar{\ell}_j),
\end{equation}
where $P_j$ is the perplexity of sentence $s_j$ under the PPL model. A larger $P_j$ indicates that the sentence is more difficult to predict from its preceding text. We therefore rank candidate sentences in descending order of $P_j$ and consider the highest-perplexity sentences first.

For a document containing $T$ tokens, we select $K$ insertion positions, where $K=\left\lfloor T/1000\right\rfloor$. The beginning of each sentence $s_j$ serves as a candidate insertion position, with the preceding text as the context and $s_j$ as the continuation. Following the perplexity ranking, the annotation model retains a position only if its continuation is meaningfully related to and inferable from its context. This process continues until the target number of positions is reached or no eligible candidates remain.

\paragraph{Annotation Generation and Refinement.}
After selecting the insertion positions, we form local context--continuation pairs around them. Let $b_1<\cdots<b_K$ be the token indices of the selected positions. We define $b_0=1$ and $b_{K+1}=T+1$ as the document boundaries. For each $i\in\{1,\ldots,K\}$, the local context and continuation are
\begin{equation}
c_i=(x_{b_{i-1}},\ldots,x_{b_i-1}), \qquad
y_i=(x_{b_i},\ldots,x_{b_{i+1}-1}).
\end{equation}
Thus, $c_i$ is the original text between the previous selected position and the current one, while $y_i$ begins with the selected sentence and ends immediately before the next selected position. For $i=1$, the context begins at the start of the document; for $i=K$, the continuation extends to the end of the document. 

For each pair $(c_i,y_i)$, the annotation model produces multiple initial book-note-style annotations. We use this format because expository notes resemble text found in natural pre-training corpora and can be inserted without an abrupt change in voice or structure. Unlike a conventional CoT trace written as first-person deliberation, each annotation uses third-person or impersonal narration. It summarizes the relevant information in $c_i$, states the missing conceptual or discourse connection, and describes why $y_i$ follows. The generation prompt asks the model to explain this dependency without restating $y_i$ and typically limits each annotation to 500--1,000 words.
The initial annotations serve as separate starting points for subsequent REER refinement and provide alternative explanations of the same transition. Before refinement, the annotation model checks whether each initial annotation satisfies the length requirement and avoids target leakage. Any annotation that fails either check is discarded.

Following REER \citep{wang2025reer}, we use the perplexity of the observed continuation $y_i$ as the optimization signal. If an annotation captures a useful dependency from $c_i$ to $y_i$, conditioning on that annotation should reduce the perplexity of $y_i$. 
Let $y_i=(y_{i,1},\ldots,y_{i,L_i})$, where $L_i=|y_i|$ is the continuation length in tokens. For any candidate annotation $a$, we define continuation perplexity as
\begin{equation}
\operatorname{PPL}_{\mathrm{ppl}}(y_i\mid c_i,a)
=\exp\left(
-\frac{1}{L_i}\sum_{q=1}^{L_i}
\log p_{\mathrm{ppl}}(y_{i,q}\mid c_i,a,y_{i,<q})
\right),
\end{equation}
where $y_{i,q}$ is the $q$-th token of $y_i$ and $y_{i,<q}=(y_{i,1},\ldots,y_{i,q-1})$ is its preceding continuation prefix. The no-annotation baseline, written as $\operatorname{PPL}_{\mathrm{ppl}}(y_i\mid c_i)$, is computed by omitting $a$ from the conditioning sequence.

Let $\mathcal{A}_i$ denote the candidate annotation space for the $i$-th context--continuation pair. The perplexity objective is
\begin{equation}
\widetilde{a}_i=\arg\min_{a\in\mathcal{A}_i}
\operatorname{PPL}_{\mathrm{ppl}}(y_i\mid c_i,a),
\end{equation}
where $\widetilde{a}_i$ denotes an annotation with minimum continuation perplexity in $\mathcal{A}_i$. A lower objective value means that the annotation makes $y_i$ easier for the PPL model to predict.
Enumerating all possible annotations is infeasible, so we perform an iterative, gradient-free search. The equations below describe the refinement trajectory of one initial annotation. The initial-candidate index is omitted for clarity. We split each annotation at paragraph boundaries into an ordered sequence of segments and refine one segment at a time. Let $a_i^{(0)}$ be the complete initial annotation before refinement, let $r\in\{0,\ldots,R_i-1\}$ index the refinement step, and let $R_i$ be the maximum number of refinement steps for pair $i$. At step $r$, the annotation model generates candidate replacements for one segment while leaving the other segments unchanged. Each replacement forms a complete candidate annotation in the finite rewrite set $\mathcal{R}_i^{(r)}$. We update the annotation by
\begin{equation}
a_i^{(r+1)}=\arg\min_{a\in\mathcal{R}_i^{(r)}\cup\{a_i^{(r)}\}}
\operatorname{PPL}_{\mathrm{ppl}}(y_i\mid c_i,a),
\end{equation}
where $a_i^{(r+1)}$ is the annotation retained for the next step. Including the current annotation $a_i^{(r)}$ in the candidate set ensures that continuation perplexity cannot increase from one refinement step to the next. Let $r_i^{\mathrm{stop}}\leq R_i$ denote the step at which refinement stops because no rewrite provides the required improvement or the step budget has been reached. The same refinement procedure is applied to every initial annotation, and $a_i^{*}$ denotes the final candidate with the lowest continuation perplexity among all resulting trajectories.

For a refined annotation $a_i^{*}$, define its perplexity reduction relative to no annotation as
\begin{equation}
\Delta_i=
\operatorname{PPL}_{\mathrm{ppl}}(y_i\mid c_i)
-\operatorname{PPL}_{\mathrm{ppl}}(y_i\mid c_i,a_i^{*}).
\end{equation}
Here, $\Delta_i>0$ means that the annotation makes the continuation easier to predict. We retain $a_i^{*}$ only if $\Delta_i>0$. Otherwise, the insertion position is discarded.

\paragraph{Augmented Corpus Construction.}

Each accepted annotation is inserted before its target continuation between the markers \texttt{<annotation\_begin>} and \texttt{<annotation\_end>}, while all source tokens retain their original order. Applying REER-PT to approximately 23B source tokens produces a 42B-token augmented corpus. The corpus remains compatible with standard next-token prediction and requires no online reasoning rollout or specialized training architecture.

\section{Experiments}

\subsection{Data Analysis}

We analyze the augmented data from two perspectives: predictability, measured by perplexity, and repetition, measured by exact 13-gram self-repetition and annotation-to-source overlap.

\paragraph{Perplexity.}
We compare three annotation conditions. \textit{No annotation} contains only the source document, \textit{Initial} inserts the filtered initial annotations before perplexity-guided refinement, and \textit{Optimized} inserts the final annotations retained after refinement and comparison with the no-annotation baseline. We evaluate PPL at three scopes. \textit{Full augmented data} includes all tokens in each condition, including annotation tokens when present. \textit{Original tokens} includes only the unchanged source tokens, although their predictions may condition on preceding annotations. \textit{Selected continuations} includes only the selected continuations $y_i$. At each scope, PPL is computed with the PPL model over the corresponding tokens. The first two scopes provide global measurements over the full data and original tokens, whereas the third provides a local measurement over the selected continuations targeted during refinement.

\begin{table}[h]
    \centering
    \footnotesize
    \caption{Perplexity comparisons for annotation insertion and refinement. The Original tokens scope excludes annotation tokens from the evaluated positions, and the Selected continuations scope includes only the selected continuations.}
    \label{tab:data_analysis}
    \begin{tabular*}{0.98\linewidth}{@{\extracolsep{\fill}}llrrr@{}}
        \toprule
        Evaluation scope & Comparison & Before PPL & After PPL & PPL reduction \\
        \midrule
        Full augmented data & No annotation $\rightarrow$ Optimized & 18.68824 & 11.40323 & \textbf{7.28501} \\
        Original tokens & No annotation $\rightarrow$ Optimized & 18.68824 & 17.65746 & \textbf{1.03078} \\
        Full augmented data & Initial $\rightarrow$ Optimized & 11.89156 & 11.40323 & \textbf{0.48833} \\
        Original tokens & Initial $\rightarrow$ Optimized & 18.07970 & 17.65746 & \textbf{0.42224} \\
        \midrule
        Selected continuations & No annotation $\rightarrow$ Optimized & 24.78340 & 20.54801 & \textbf{4.23539} \\
        Selected continuations & Initial $\rightarrow$ Optimized & 21.93265 & 20.54801 & \textbf{1.38464} \\
        \bottomrule
    \end{tabular*}
\end{table}

\begin{figure}[h]
    \centering
    \includegraphics[width=\linewidth]{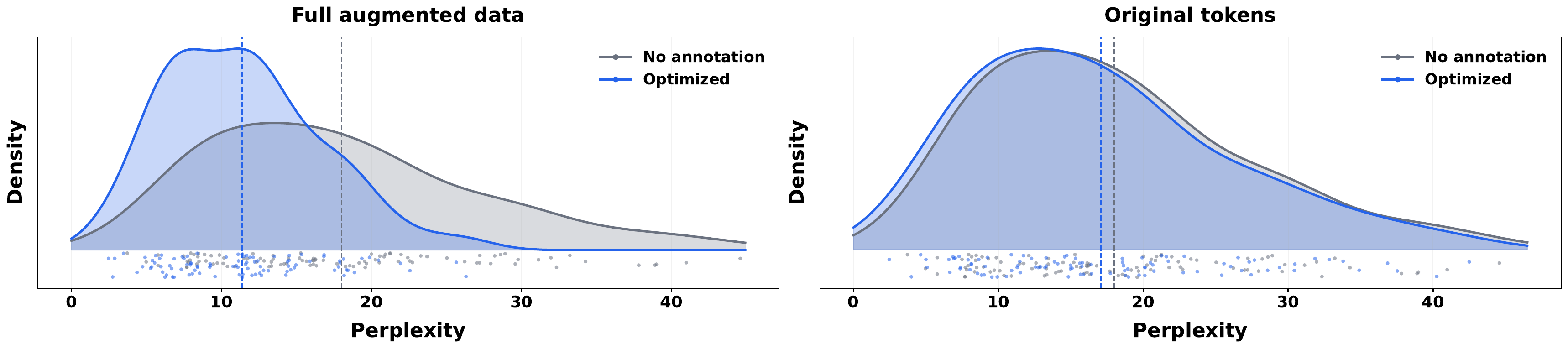}
    \caption{PPL distributions under different annotation conditions and evaluation scopes.}
    \label{fig:distribution_overview}
\end{figure}

As shown in Table~\ref{tab:data_analysis}, every comparison yields a positive PPL reduction at both the global and local scopes. At the global scope, the optimized condition reduces full augmented-data PPL by 7.28501 and original-token PPL by 1.03078 relative to no annotation. The larger reduction on the full augmented data may partly arise because the model-generated annotations are more fluent and predictable than the source text. Importantly, PPL also decreases on the original tokens, showing that the improvement extends to the unchanged source content rather than being confined to the inserted annotations. Relative to the initial condition, perplexity-guided refinement further reduces full augmented-data PPL by 0.48833 and original-token PPL by 0.42224. At the local scope, optimized annotations reduce selected-continuation PPL by 4.23539 relative to no annotation and by 1.38464 relative to the initial annotations. Together, these results support the effectiveness of both annotation insertion and perplexity-guided refinement. Figure~\ref{fig:distribution_overview} shows the corresponding PPL distributions. Each panel displays smoothed distributions and 100 sampled observations.

\paragraph{Repetition.}
We next measure repetition within each text and exact copying from a source document into its annotations. We use 13-grams throughout. By default, we construct word 13-grams after splitting text on whitespace. If whitespace splitting produces fewer than two words, as can occur in text without whitespace-separated words such as Chinese, we instead construct character 13-grams after removing whitespace. We extract overlapping 13-grams with a sliding window and exclude text containing fewer than 13 words or characters under the applicable representation from the corresponding mean.

Let $\mathcal{G}_{13}(z)$ denote the multiset of all 13-gram occurrences in text $z$, and let $\operatorname{uniq}(\mathcal{G}_{13}(z))$ denote the set of distinct 13-grams in that multiset. For any text $z$, we define the self-repetition ratio as
\begin{equation}
    R_{\mathrm{self}}(z)
    = \frac{|\mathcal{G}_{13}(z)| - |\operatorname{uniq}(\mathcal{G}_{13}(z))|}
    {|\mathcal{G}_{13}(z)|}.
\end{equation}
Here, $|\mathcal{G}_{13}(z)|$ counts 13-gram occurrences with multiplicity, so $R_{\mathrm{self}}(z)$ is the fraction of occurrences beyond the first occurrence of each distinct 13-gram. For a final annotation $a$ and its source document $x$, we define the directional annotation-to-source overlap as
\begin{equation}
    R_{\mathrm{cross}}(a,x)
    = \frac{\sum_{g \in \mathcal{G}_{13}(a)} \mathbf{1}[g \in \mathcal{G}_{13}(x)]}
    {|\mathcal{G}_{13}(a)|},
\end{equation}
where $g$ denotes an annotation 13-gram occurrence and $\mathbf{1}[\cdot]$ is the indicator function, equal to 1 when its condition is true and 0 otherwise. Because the denominator counts annotation 13-grams, $R_{\mathrm{cross}}(a,x)$ measures the fraction of annotation 13-gram occurrences that also appear in the source. It is not a source-to-annotation coverage measure. Table~\ref{tab:ngram_repetition} reports arithmetic means of the relevant per-document or per-annotation ratios.

\begin{table}[h]
    \centering
    \small
    \caption{Mean exact 13-gram repetition and annotation-to-source overlap. Lower values indicate less repetition or exact copying.}
    \label{tab:ngram_repetition}
    \begin{tabular*}{0.68\linewidth}{@{\extracolsep{\fill}}lr@{}}
        \toprule
        Metric & Mean percentage \\
        \midrule
        Source-document self-repetition & 0.615\% \\
        Annotation self-repetition & 0.203\% \\
        Annotation-to-source exact overlap & 0.051\% \\
        \bottomrule
    \end{tabular*}
\end{table}

As shown in Table~\ref{tab:ngram_repetition}, annotations have a lower mean self-repetition ratio than source documents, at 0.203\% versus 0.615\%. The mean annotation-to-source exact-overlap ratio is 0.051\%, indicating that only a small fraction of annotation 13-gram occurrences exactly match a source span. These measurements show little repetition or verbatim copying.

\subsection{Pre-training Experiments}

\paragraph{Setup.}
We construct two pre-training mixtures. The \textit{raw mixture} combines the 23B-token source corpus with a 500B-token general pre-training corpus, for approximately 523B tokens in total. The \textit{REER-PT mixture} replaces the source corpus with its 42B-token augmented version and uses the same 500B-token general corpus, for approximately 542B tokens in total. We train a 680M-parameter language model from scratch on each mixture. We refer to the model trained on the raw mixture as the \textit{raw baseline} and the model trained on the augmented mixture as the \textit{augmented-data model}. The two runs use the same architecture, tokenizer, optimizer configuration, and other training hyperparameters. 
Their corpus composition and total token counts differ because REER-PT adds annotation tokens. This difference is part of the data transformation being evaluated.

\begin{figure}[h]
    \centering
    \includegraphics[width=\linewidth]{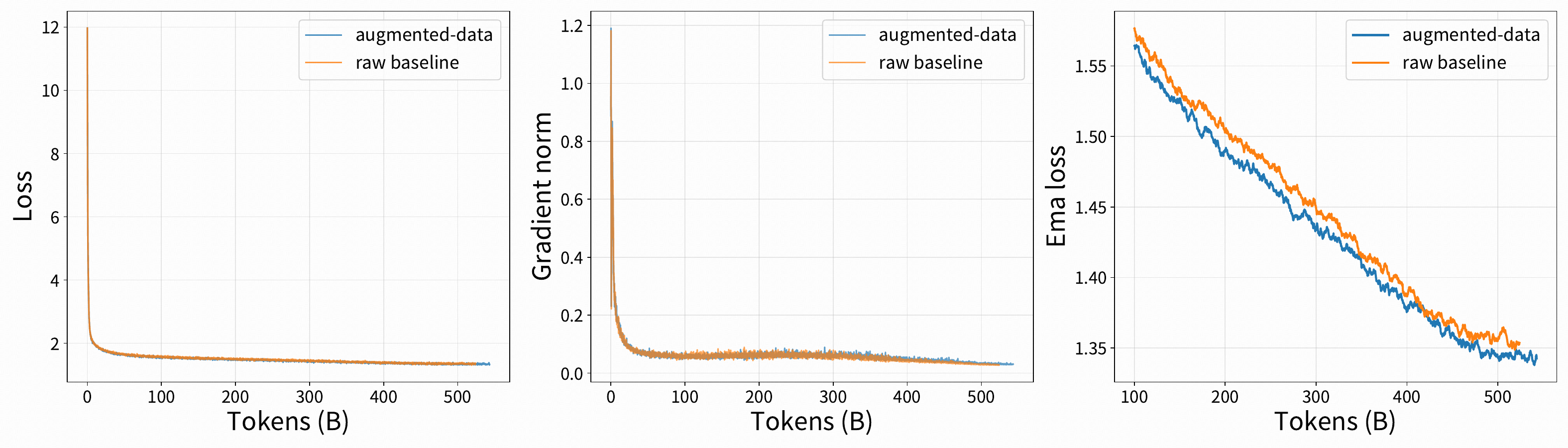}
    \caption{Training dynamics of the raw baseline and augmented-data model: raw training loss (left), gradient norm (center), and EMA-smoothed training loss after 100B consumed tokens with decay $0.95$ (right).}
    \label{fig:training_dynamics}
\end{figure}

\paragraph{Training dynamics.}
Figure~\ref{fig:training_dynamics} compares the optimization trajectories of the raw baseline and the augmented-data model. The left and center panels show raw training loss and gradient norm over the complete runs. The gradient-norm trajectories are similar, while the augmented-data model generally reaches lower training loss later in training. The right panel reports the EMA-smoothed training loss after 100B consumed tokens,
\begin{equation}
\widetilde{L}_t=0.95\widetilde{L}_{t-1}+0.05L_t,
\end{equation}
where $t$ indexes training steps, $L_t$ is the raw training loss at step $t$, and $\widetilde{L}_t$ is the smoothed loss, initialized with $\widetilde{L}_1=L_1$. The smoothed augmented-data-model curve generally remains below the raw-baseline curve later in training.

\begin{table}[H]
    \centering
    \footnotesize
    \caption{Evaluation of the raw baseline and augmented-data model on public benchmarks. Scores are reported on a 0--100 scale, and higher values are better. For each benchmark, $\Delta$ is the augmented-data-model score minus the raw-baseline score, measured in percentage points.}
    \label{tab:pretrain_evaluation}
    \begin{tabular*}{0.94\linewidth}{@{\extracolsep{\fill}}lrrr@{}}
        \toprule
        Benchmark & Augmented-data model & Raw baseline & $\Delta$ \\
        \midrule
        \multicolumn{4}{@{}l}{\textit{Knowledge}} \\
        C-Eval \citep{huang2023ceval} & 77.19 & 76.67 & +0.52 \\
        MMLU-Pro \citep{wang2024mmlupro} & 36.85 & 35.95 & \textbf{+0.90} \\
        SuperGPQA \citep{mapteam2025supergpqa} & 23.15 & 22.55 & +0.60 \\
        Chinese SimpleQA \citep{he2024chinesesimpleqa} & 34.24 & 33.71 & +0.53 \\
        \addlinespace
        \multicolumn{4}{@{}l}{\textit{General Reasoning}} \\
        DROP \citep{dua2019drop} & 41.51 & 40.11 & \textbf{+1.40} \\
        BBH \citep{suzgun2022bbh} & 55.06 & 52.99 & \textbf{+2.07} \\
        ZebraLogic \citep{lin2025zebralogic} & 6.73 & 6.20 & +0.53 \\
        ProcBench \citep{fujisawa2024procbench} & 5.93 & 5.57 & +0.36 \\
        \addlinespace
        \multicolumn{4}{@{}l}{\textit{STEM Reasoning}} \\
        GPQA-Diamond \citep{rein2023gpqa} & 27.88 & 25.81 & \textbf{+2.07} \\
        MATH \citep{hendrycks2021math} & 35.70 & 34.20 & \textbf{+1.50} \\
        OlympiadBench \citep{he2024olympiadbench} & 13.19 & 11.70 & \textbf{+1.49} \\
        \addlinespace
        \multicolumn{4}{@{}l}{\textit{Code}} \\
        MBPP+ \citep{liu2023evalplus} & 58.20 & 60.85 & $-2.65$ \\
        HumanEval+ \citep{liu2023evalplus} & 58.54 & 60.37 & $-1.83$ \\
        LiveCodeBench \citep{jain2024livecodebench} & 3.94 & 5.73 & $-1.79$ \\
        \bottomrule
    \end{tabular*}
\end{table}

\paragraph{Evaluation.}
We evaluate both pre-trained models in four categories. The knowledge category includes C-Eval \citep{huang2023ceval}, MMLU-Pro \citep{wang2024mmlupro}, SuperGPQA \citep{mapteam2025supergpqa}, and Chinese SimpleQA \citep{he2024chinesesimpleqa}. The general-reasoning category includes DROP \citep{dua2019drop}, BBH \citep{suzgun2022bbh}, ZebraLogic \citep{lin2025zebralogic}, and ProcBench \citep{fujisawa2024procbench}. The STEM-reasoning category includes GPQA-Diamond \citep{rein2023gpqa}, MATH \citep{hendrycks2021math}, and OlympiadBench \citep{he2024olympiadbench}. The code-generation category includes HumanEval+ and MBPP+ from EvalPlus \citep{liu2023evalplus}, together with LiveCodeBench \citep{jain2024livecodebench}.

As shown in Table~\ref{tab:pretrain_evaluation}, positive gains occur across the knowledge, general-reasoning, and STEM-reasoning categories. BBH and GPQA-Diamond each improve by 2.07 percentage points, followed by MATH, OlympiadBench, and DROP with gains of 1.50, 1.49, and 1.40 points, respectively. MMLU-Pro improves by 0.90 points, while C-Eval, SuperGPQA, Chinese SimpleQA, ZebraLogic, and ProcBench show smaller gains of 0.36--0.60 points.
However, all three code-generation benchmarks decline. The changes are $-2.65$ points on MBPP+, $-1.83$ points on HumanEval+, and $-1.79$ points on LiveCodeBench. Our case-level analysis suggests that inserting natural-language annotations into code documents can disrupt local program structure and encourage models to mix explanatory text with executable code. Such outputs may express a plausible solution but fail benchmarks that require concise, syntactically valid programs. These results motivate future work on reasoning-annotation formats tailored to code documents.

\section{Conclusion}

We introduced REER-PT, an offline framework that augments raw pre-training data with concise reasoning annotations. REER-PT selects difficult continuations that remain inferable from context and retains annotations that reduce continuation perplexity while satisfying length and target-leakage constraints. The resulting corpus preserves all source tokens and remains compatible with standard next-token prediction. Across the full augmented data, original tokens, and selected continuations, the reported PPL reductions range from 0.42 to 7.29, while the mean exact annotation-to-source 13-gram overlap is only 0.05\%. In controlled pre-training experiments with 680M-parameter models, the augmented-data model improves several knowledge and reasoning benchmarks by up to 2.07 percentage points. Together, these results suggest that sparse, perplexity-guided reasoning augmentation can improve continuation predictability and downstream capabilities.

\section{Limitations and Future Work}

Our experiments use a 680M-parameter model and a single pre-training recipe, so the behavior of REER-PT at larger model and data scales remains unknown. Data construction also depends on both model choices. A different PPL model may assign different perplexities, select different insertion positions, or provide different refinement signals, while a different annotation model may make different inferability judgments or produce a different annotation style. In addition, the current book-note-style format is designed primarily for natural-language documents. In code documents, natural-language insertions can disrupt program structure and encourage explanatory text inside generated programs, consistent with the observed code-generation regressions. Future work should study larger-scale training, annotation density, and the choices of PPL and annotation models. It should also develop structure-aware annotation formats for code and other specialized domains and identify which types of implicit dependency benefit most from explicit reasoning annotations.

\clearpage

\bibliographystyle{unsrtnat}
\bibliography{main}

\end{document}